**Title**

# A primer on evaluation methods for large language models in healthcare

**Authors**

Suzannah E McKinney, MBChB[1]

Phuc Vu, MPH[1]

Samuel A Justice, PhD[1]

Christopher Humphries, MBChB[2,3]

Alyssa Pradhan, MBBS[4]

Timothy J Keyes, PhD[5]

Kameron C Black, DO[6]

John L Havlik, MD MBA[7]

David JH Wu, MD[8]

Jack W O'Sullivan, MBBS DPhil[5,6]

Ethan Goh, MD[9]

Sarah F Mercaldo, PhD[1,10,11]

James M Hillis, MBBS, DPhil[1,11,12]

**Affiliations**

[1]Mass General Brigham AI, Mass General Brigham, Boston, MA, USA

[2]Generative AI Laboratory, School of Informatics, University of Edinburgh, Edinburgh, UK

[3]University of Edinburgh Centre for Cardiovascular Science, Edinburgh, UK

[4]Experimental Medicine Division, Nuffield Department of Medicine, University of Oxford, Oxford, UK

[5]Department of Biomedical Data Science, Stanford University School of Medicine, Stanford, CA, USA

[6]Department of Medicine, Stanford University School of Medicine, Stanford, CA, USA

[7]Department of Psychiatry, Stanford University School of Medicine, Stanford, CA, USA

[8]Department of Radiation Oncology, Stanford University School of Medicine, Stanford, CA, USA

[9]Stanford Division of Computational Medicine, Stanford University, Stanford, CA, USA

[10]Department of Radiology, Mass General Brigham, Boston, MA, USA

[11]Harvard Medical School, Boston, MA, USA

[12]Department of Neurology, Mass General Brigham, Boston, MA, USA

**Word count:** 3999 (introduction to conclusion excluding figure / table legends)

**Figure count:** 4 (included at the end of each section)

**Table count:** 1 (included at the end of each section)

**Date of revision:** October 14 2026

**Abstract**

Large language models (LLMs) have a growing range of applications in medicine, and their evaluation is critical for ensuring they provide benefit and not harm. This evaluation can be more challenging than traditional machine learning for many reasons, including probabilistic and open-ended outputs, and behavior that shifts with prompt design and accumulated context. This review covers four key areas of LLM evaluation: principles of study design, statistical methods, capability evaluation and clinical application evaluation. Capability evaluation considers different benchmarks, including multiple-choice, agentic and multi-turn benchmarks, alongside operational metrics like token usage and latency. Clinical application evaluation addresses establishing accuracy of free text outputs, such as human review and LLM-as-a-judge, and clinical trial approaches. Across sections, we describe underlying concepts and potential pitfalls, while emphasizing the importance of aligning evaluation methods with the research question. Together, this article aims to provide a pragmatic basis for designing and executing rigorous evaluations of healthcare LLMs.

## Introduction

The emergence of large language models (LLMs) in medicine has enabled a range of clinical and research uses.[1] They include clinical reasoning with physician-level performance,[2] agents that complete tasks like medical record reviews,[3,4] and chatbots that converse directly with patients.[5] These applications may involve models built specifically for healthcare or broader general-purpose models applied in the healthcare setting,[6] and access to them may be open to the public or restricted to healthcare personnel.

LLMs have more varied outputs and capabilities than traditional machine learning (ML) models, including free text, multi-turn dialogue and agentic task completion. The underlying task of evaluating them often returns to familiar items. What is the research question? What is an appropriate reference standard? Is the study adequately powered?

This review aims to provide a pragmatic introduction to the evaluation of LLMs (Figure 1). It begins by considering study design principles and common statistical methods. It then describes capability evaluations including the use of benchmarks and operational metrics. It later addresses clinical application evaluations including establishing accuracy of free text outputs and clinical trial approaches. This review refers to models and LLMs interchangeably. It also uses these terms for simplicity while recognizing LLMs are often part of a broader architecture that could include agent harnesses for facilitating agentic functions or context engineering for determining which data a model receives.[7,8]

**Figure 1:** From research question to statistical analysis plan: a framework for evaluating healthcare LLMs. Figure created with ChatGPT (GPT-5.6 Luna; image generation, default settings); accessed October 6, 2026.

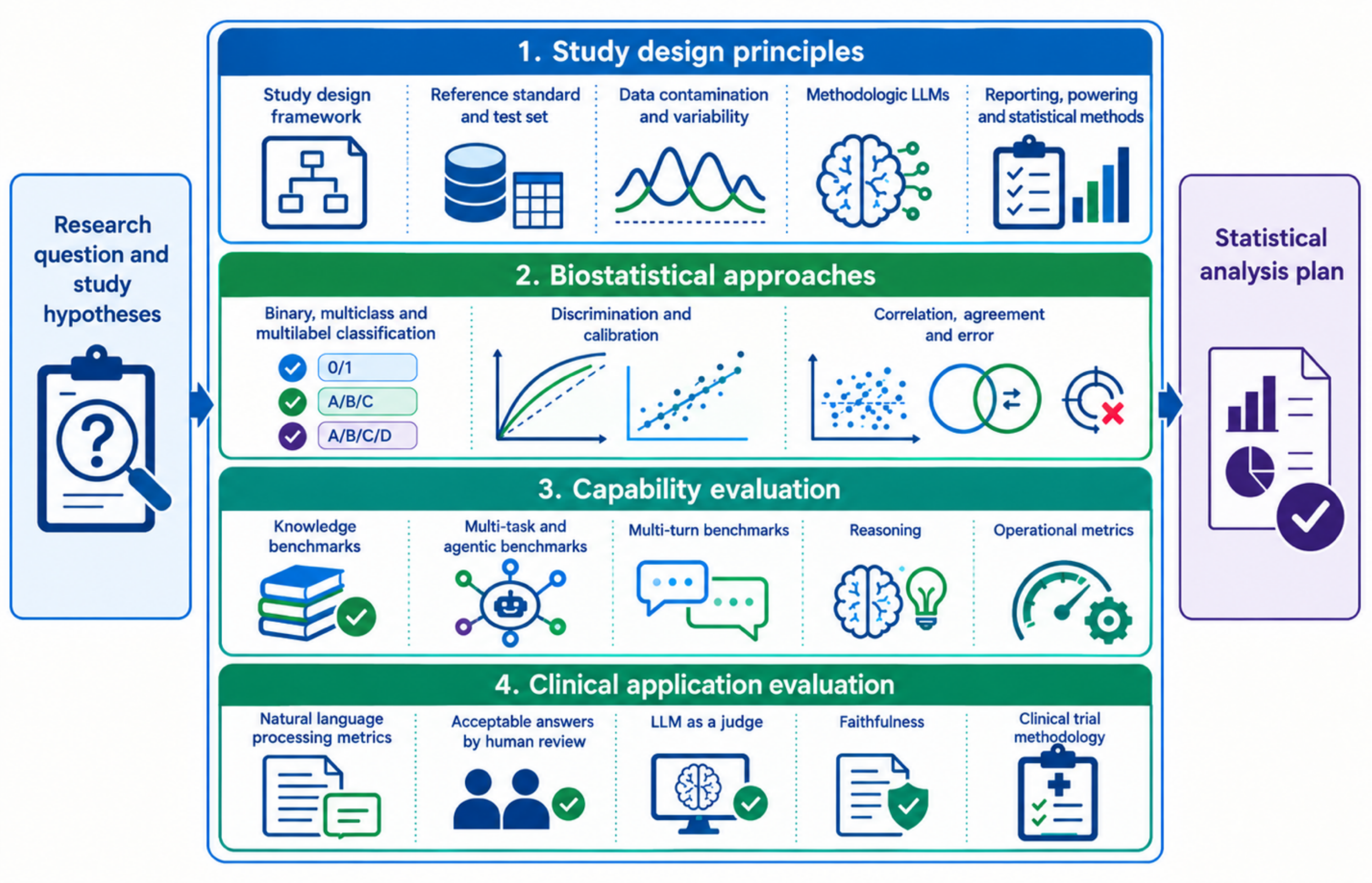

## Study design principles

### Study design framework

LLM evaluation is like most medical research in that it requires an objective with associated endpoints that translate to null and alternative hypotheses. These elements are ideally prespecified in a study protocol or statistical analysis plan. The assessments may aim to demonstrate minimum performance, compare models with one another, or compare a model back to a human clinician. Predetermined thresholds may be set, against which performance passes or fails. Assessments may also provide descriptive analyses about operational metrics like compute usage, latency and cost.

### Reference standard

A reference standard, often known as the ground truth, is determined by the research question. For task-specific evaluations, it may be a discrete value such as the presence or absence of a diagnosis in an LLM output. It can also involve a benchmark as described below. Such standards vary widely in their rigor including utilizing already codified values like ICD-10 codes in patient charts or obtaining consensus opinions from expert panels.[9] For free-text generation, there may not be a single correct output. An evaluation may therefore rely on structured human or LLM-based assessment across multiple criteria in a rubric structure. It may also compare outputs with an exemplar or gold standard such as a radiologist-authored report. Human performance, including inter-clinician variability, is frequently an important comparator.

### Training and testing datasets

While traditional ML models commonly use highly curated training datasets, LLMs are often trained on large, undisclosed corpora. The components of the broader architecture, like agent harnesses, may require additional development and their own training datasets. It helps to know this structure and training approach when designing evaluations.

Two key concerns for testing dataset selection are data contamination and generalizability. The former can manifest if testing data were part of training corpora; published case reports might otherwise provide a convenient, well-structured narrative to test performance.[10] Benchmarks, as discussed below, aim to avoid contamination although may become part of training corpora once publicly released. Additionally, an evaluation rubric should only provide data available at the relevant point in the clinical pathway to avoid temporal leakage. They should not inadvertently include downstream diagnostic or treatment steps such as antibiotic administration when evaluating a model that identifies sepsis.

Generalizability involves ensuring that LLMs work across all intended settings. Especially if an LLM or its broader architecture were developed at a single hospital or healthcare system, a concern exists for overfitting to the original site and performance should be evaluated at other sites. Similarly, the demographic representativeness of a test dataset warrants consideration; prespecified subgroup analyses should typically be performed for age group, sex, race and ethnicity to detect performance disparities that aggregate metrics otherwise obscure. Representation of real-world spectrum of disease is also important given clinical practice involves equivocal presentations rather than only textbook presentations. For studies involving actual clinical cases, consecutive case selection can help achieve this spectrum.

<u>Model stability over time</u>

A shift in training data can occur when models are updated with data from user activity. As mitigation, evaluations can specify a locked model version rather than a continuously updated model.[11] Furthermore, given LLMs can be influenced by information retained within a session, evaluations may use stateless single-turn application programming interface (API) calls to avoid earlier cases influencing later outputs.[12] Repeated testing can also be used to check for response evolution.

### Run-to-run variability

LLM evaluation faces a challenge from stochastic output: the same input can produce different outputs, complicating comparison within and across models. Mitigation strategies include conducting repeated runs for each case and pairing analyses on the same cases when comparing models. The phrasing of a prompt can also impact performance. Additionally, evaluations can utilize advanced features available via API access, including sampling or temperature settings, to minimize output randomness; they should be fixed across runs unless they are specifically being tuned.[13] A fixed random seed, if supported, can further reduce residual randomness across runs.

### Using LLMs for the evaluation

Beyond LLMs being the subject of evaluations, they can help perform evaluations, referred to here as a methodologic LLM. For example, an LLM could assist case selection by reviewing records against inclusion criteria as already occurs in clinical trials.[14,15] An LLM used to make evaluation decisions, known as "LLM-as-a-judge", is discussed more below. To ensure fairness and consistency, multiple methodologic LLMs can perform the same task with manual review performed on a sample of outputs.[16] Prompts provided to a methodologic LLM should be unambiguous and reported in a publication.

### Reporting checklists

Reporting guidelines have become a requirement for many journals to ensure manuscripts include critical elements. They improve comparability, reproducibility and appraisal, and can serve as a design-stage checklist. TRIPOD-LLM provides a checklist of items for different LLM study designs and tasks, emphasizing transparency and task-specific performance reporting.[11] Discipline-specific instruments include FLAIR for radiology.[17]

Sample size and powering

The number of cases needed in a study depends on the research question and expected degree of results variation. In practice, case availability, time and funding can also impose constraints; a need for human review, which journals and reviewers often expect, can drive these factors. Ideally studies comparing systems should be powered a priori through a sample-size calculation that determines the number of cases needed to detect a meaningful or clinically important difference. With the rapidly evolving nature of LLM research, the acceptable thresholds for many metrics are still being determined.

Confidence, significance and meaning

Statistical methods should be appropriate for the research question and the type of data collected. While results are often reported using p-values, a threshold of $p<0.05$ does not establish that a difference is real, clinically meaningful, or large; a very small difference can be statistically significant in a large sample. It is often desirable to report confidence intervals that help readers understand both the possible size of an effect and the uncertainty around it: for 95% confidence intervals, if repeated studies were conducted in the same way, approximately 95% of the intervals will contain the true value. A common method for calculating these intervals is bootstrapping that involves resampling within a dataset while repeatedly calculating the metric.[18] If data will be reused, including asking multiple questions about a single input or a single patient, multiple testing should be accounted for.[19] There are also alternative statistical paradigms to the standard frequentist methods that this review focuses on.[20] Beyond individual evaluations, tiered evidence frameworks situate findings in a broader hierarchy of evidence:

from retrospective benchmarking through silent prospective deployment to randomized comparison against routine care.[21,22]

## Statistical methods

Many traditional statistical methods are used for LLM evaluations given LLM outputs are often reduced to categorical or numeric forms (Table 1).

### Categorical outputs

Classification tasks involve categorical outputs and include several types. Binary classification involves picking one of two options, like presence or absence of a condition. Multiclass and ordinal classification involve picking one of three or more options, such as assigning disease severity grade. Multilabel classification involves potentially picking more than one option from a list, like generating problem list diagnoses. Beyond returning the respective labels for classification tasks, an LLM may provide an underlying probability or confidence score for labels. It is commonly a numeric value between 0 and 1 and allows assessment of model discrimination for how well it ranks cases. If the score reflects a true probability that intends to match frequency (i.e., 0.8 corresponds to 80% occurrence), calibration can be assessed.

For binary tasks, standard metrics include sensitivity, specificity, positive and negative predictive value, and accuracy.[23,24] Accuracy alone can be misleading when one class is rare:[25] if only 5% of notes document a condition, a model that always answers "negative" reaches 95% accuracy while missing every positive case. Balanced accuracy,[26] which averages sensitivity and specificity, and the F1 score,[27] which rewards true positive outputs without too many false positive outputs, are useful in this setting.

For multiclass tasks, raw agreement does not correct for chance and can mislead with imbalanced classes. Cohen's kappa measures agreement while correcting for chance.[28] For ordinal categories, where disagreement between certain categories equates to a larger error, weighted kappa is useful.[29] Per-class performance can be summarized by macro-averaging, which weights every category equally, or micro-averaging, which weights every case equally; the two diverge under class imbalance and the choice should be reported.

For multilabel tasks, a key question is how much outputs overlap a reference set. The Dice coefficient[30] and Jaccard index[31] (intersection over union) both measure overlap and differ in normalization. The Tversky index[32] weights incorrect additions and missed items differently; it helps when one error is more costly such as omitting a diagnosis rather than adding one.

Discrimination may be thought of as each score reflecting a threshold that yields its own sensitivity and specificity. Varying these thresholds and plotting sensitivity against the false-positive rate (1 − specificity) creates a receiver operating characteristic (ROC) curve.[33] The area under a ROC curve (AUC) shows the probability that the model gives a randomly chosen positive case a higher score than a randomly chosen negative one: 0.5 is chance, 1 is perfect. If there is a negative class imbalance and the impact of false positive cases needs to be considered, area under the precision-recall curve (AUPRC) can provide an informative metric by plotting precision (positive predictive value) against recall (true positive rate).[34] If a model only provides end-users with a binary output, the binary classification metrics should take precedence to discrimination metrics and the operating point threshold should be prespecified.

Calibration matters when a probability is shown to a clinician or determines downstream clinical steps, especially as studies have shown LLMs to be over-confident.[35] The Brier score gives a single summary number, though it blends calibration with discrimination rather than isolating either.[36] Decision curve analysis calculates the net benefit of a model guiding clinical decisions across different threshold probabilities, and compares this benefit to treating all patients or none. It is most useful when an output results directly in a clinical action.[37]

Beyond categorical model outputs, LLM evaluations often use rubrics or surveys featuring binary or multiclass labels including Likert scales. These results are commonly displayed as frequencies alongside statistical tests that evaluate for significance in comparisons being made. McNemar's test compares paired dichotomous data including the concordance of two binary outputs.[38] Fisher's exact test and Chi-square tests compare counts or frequencies like the number of correct and incorrect responses between different models; the former is more appropriate for small sample sizes.[39] The Wilcoxon rank sum or signed rank tests, and advanced techniques like ordinal mixed-effects models, can compare multiclass labels with an ordinal nature like Likert scales.[40,41]

Numeric outputs

Correlation, agreement, and error each answer different questions about numeric outputs and should be reported separately. Correlation asks whether predicted and reference values vary together: Pearson for a linear relationship, Spearman for a monotonic one.[42,43] Correlation does not incorporate agreement, which asks whether values coincide on the same scale. Agreement can instead be assessed with intraclass correlation coefficient and concordance correlation coefficient.[44,45] Error quantifies the typical deviation. The root mean square error and mean absolute error report error in the units of the output; the former penalizes larger misses more heavily.

**Table 1:** Statistical techniques

| Evaluation Category | Metric | Purpose | Example Application (Healthcare LLM) |
|---|---|---|---|
| **Binary classification** | Sensitivity (recall, true-positive rate), specificity (true-negative rate), positive predictive value (precision), negative predictive value, accuracy, balanced accuracy, F1 score | Measure classification performance given predicted labels | Presence/absence detection (e.g., condition documentation, abnormal-finding flagging) |
| **Multiclass / ordinal agreement** | Cohen κ, weighted κ (linear or quadratic), macro- and micro-averaged precision / recall / F1 | Quantify chance-corrected agreement and per-class performance | Severity grading (e.g., staging, triage category) |
| **Multilabel classification** | Dice coefficient, Jaccard index (intersection over union), Tversky index | Measure the similarity between a predicted set of items and a reference set | Set extraction and coding (e.g., ICD-10 code assignment, problem-list extraction) |
| **Discrimination and ranking** | Receiver operating characteristic (ROC) curve, area under the ROC curve (AUC), area under the precision–recall curve (AUPRC) | Evaluate overall ability to separate the classes, aggregated over a range of thresholds | Risk stratification and triage (e.g., ranking cases by predicted probability) |
| **Calibration (if model outputs probabilities)** | Brier score, calibration curve | Assess how closely predicted probabilities match observed event rates | Probability-based risk estimation (e.g., calibrated diagnosis or event probabilities) |
| **Clinical utility** | Decision curve analysis (DCA) | Plots net benefit of a decision against threshold, a minimum probability of a disease where a decision-maker chooses to take action | Decision-support tools (e.g., biopsy) |
| **Rubric or survey comparison** | McNemar's test, Fisher's exact test, Chi-square test, Rank-sum, Signed-rank, ordinal mixed-effects model | Compare binary or multiclass labels (e.g., between reference and model, or between different models) | Performance for different LLMs when interpreting MCQ questions or when being assessed with Likert scales. |
| **Correlation** | Pearson correlation, Spearman correlation | Measure linear or monotonic association between predicted and reference values | Score ranking and trend agreement (e.g., risk-score ordering vs. reference) |
| **Agreement** | Intraclass correlation coefficient (ICC), concordance correlation coefficient (CCC) | Assess whether predicted and reference values coincide on the same scale | When predicted numeric values should agree closely with a reference value (e.g., ejection fraction, tumor diameter) |
| **Error** | Root mean square error (RMSE), mean absolute error (MAE) | RMSE/MAE give error in the outcome's units | Numeric extraction and measurement (e.g., lab values, dosages, size/volume quantification) |

## Capability evaluation

Capability evaluation asks whether models possess the broad clinical knowledge, reasoning, and task-completion abilities required for use in healthcare.

### Benchmarks

The term 'benchmark' refers to a fixed dataset of tasks and associated scoring procedure (Figure 2). Benchmarks commonly have predefined correct answers that may be determined using the Delphi technique;[46] there may also be a process for reconciling novel responses. The scoring procedure draws on the same methods used to judge any model output against a reference as described in other sections. Critically, the scoring procedure applies these methods through a fixed, predetermined protocol. Benchmarks facilitate easy model comparison.

#### *Multiple-choice and knowledge benchmarks*

The initial benchmarks for evaluating LLM performance in healthcare contexts were knowledge-oriented and included MedQA[47], a set of US Medical Licensing Examination-style multiple-choice questions. Such benchmarks can provide an early signal for model performance but are increasingly considered insufficient for testing clinical readiness, capturing factual recall more than real-world utility.[48,49] They also often focus on a single timepoint rather than a multistep, iterative clinical workflow.

#### *Multi-task and agentic benchmarks*

Multi-task frameworks provide a more robust approach to knowledge benchmarking. The value of these frameworks is breadth: they characterize performance across different tasks and domains. MedHELM assesses models against a clinician-validated taxonomy of 121 real clinical tasks spanning decision support, note generation, patient communication, research and administration.[50] Limitations of these frameworks include aggregate scores obscuring task-level variation and constituent tasks not resembling the entire clinical workflow.

AI agents involve systems that autonomously plan and execute a sequence of actions, including using tools or external data sources. They therefore require evaluation frameworks that assess both quality of individual outputs and an agent's ability to navigate complex environments; for example, obtaining information via Fast Healthcare Interoperability Resources (FHIR) calls. Relevant agentic benchmarks include Medical Intelligence for Reasoning and Action (MIRA), MedAgentBench, PhysicianBench and HealthAgentBench.[3,51–53]

*Multi-turn benchmarks*

A multi-turn conversation involves a model producing a series of exchanges with users, as a chatbot would. General frameworks commonly use an LLM-as-a-judge that scores responses to a fixed set of benchmark conversations against criteria including factual accuracy and communication quality.[4,54–57] To mimic the stochastic nature of real conversations, other benchmark designs use patient actors to generate the dialogue or LLM agents prompted with a predefined case vignette.[58] These simulated users typically require their own separate validation.

*Reasoning benchmarks*

Typical methods of scoring LLM outputs assume that an output is correct if it matches a reference answer, irrespective of its reasoning. Reasoning assessment evaluates the process rather than an endpoint. This distinction matters because the black box nature of LLMs can limit their reliability and trustworthiness. Models can be prompted to divide tasks into smaller steps to demonstrate their step-by-step reasoning, known as Chain-of-Thought (CoT) reasoning.[59,60]

Multiple benchmarks exist where clinical reasoning is present in the referenced answers, including MedThink-Bench,[61] MedR-Bench,[62] Open-XDDx[63] and ER-Reason.[64] These benchmarks vary in whether reference reasoning is expert-authored or machine-generated. Separately, script concordance testing investigates how added data adjust outputs, aiming to reflect the often uncertain and dynamic nature of clinical decision making.[65]

While generated CoT reasoning can provide insight into how a model reaches its answer, several studies have shown divergence between the stated reasoning and the process driving a model's output. One study found that truncating or inserting errors into the reasoning chain often leaves the final answer unchanged, indicating that the intervening steps were not load-bearing.[66] Another study introduced input features aiming to bias model answers: models frequently produced CoT explanations that justified an incorrect answer and did not mention the biasing features in those explanations.[67] Parametric methods also allow CoT assessments; they involve erasing information in model weights about individual reasoning steps and testing whether predictions consequently change.[68]

These methods commonly depend on model-stated CoTs and a future, though still experimental, direction for reasoning assessments utilizes separate insights. The J-space by Anthropic considers a set of internal representations identified by an interpretability method called the Jacobian lens.[69] It surfaces intermediate assessments a model has formed but not verbalized. Further methods include probing classification[70] and activation patching.[71]

### Operational metrics

Operational metrics like token usage and latency are critical to consider alongside accuracy, especially as there can be a tradeoff (Figure 3).[72] Furthermore, energy footprint and sustainability have drawn increasing attention as generative AI is deployed at scale.[73]

Tokens are defined as the basic piece of text or data that an AI model receives as an input, such as a word, part of a word or punctuation.[74] For example, “Unreal!” may contain three tokens of “Un”, “real” and “!”. Token usage is the count of input and output tokens per query. Cost per query translates this usage into dollar terms, which allows health systems to budget for LLM-based tool deployment. Latency is the total time delay between submitting a prompt and receiving an answer. It can determine effective integration for real-time clinical workflows such as point-of-care decision support. It can also vary with time of day based on periods of higher usage. Broader system measures for reliability and scalability include the frequency of system outages or downtime, error and failure rates, and behavior under load.[75] Similarly, the human time needed to work alongside a model and the cost effectiveness for downstream actions from an LLM should be considered.[76,77]

Other operational characteristics may be specified to a model, including the maximum output tokens that determines the length of responses. Two additional modifiable parameters are temperature and Top-P or Top-K.[78] Temperature adjusts the randomness in generating an output: a lower temperature provides more deterministic outputs and is usually necessary for healthcare applications, while a higher temperature helps with tasks that benefit from creativity like writing fiction. Top-P reflects the underlying LLM methodology of probabilistically choosing the next token when generating responses: a lower Top-P constrains these options to a cumulative probability (e.g., the top 90% of response probabilities) and therefore provides more predictable outputs. Top-K is similar although uses a fixed number cutoff (e.g., top 3 responses). Given temperature, Top-P and Top-K influence outputs in similar ways; typically, only one is adjusted.

**Figure 2: Benchmarks.** Common types of benchmarks used for evaluation of large language models in healthcare including multiple-choice and knowledge benchmarks (A), multi-task and agentic benchmarks (B), multi-turn benchmarks (C) and reasoning benchmarks (D). Acronym: EHR: electronic health record. Figure created in BioRender (https://BioRender.com/36ygayy).

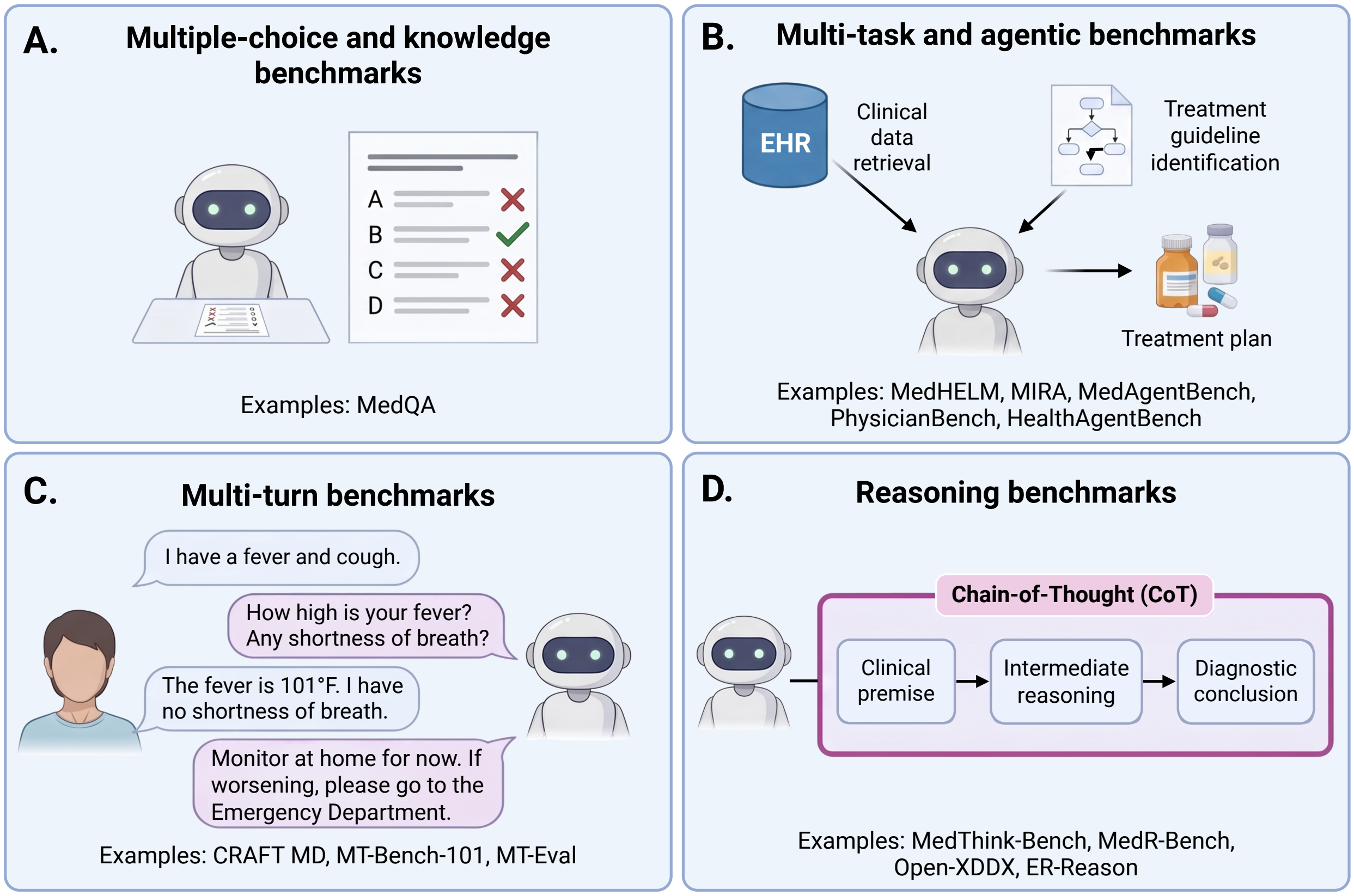

**Figure 3: Operational metrics.** Operational metrics (A) that can be used to evaluate a large language model together with operational specifications (B) that can be chosen for the large language model. Figure created with BioRender (https://BioRender.com/dugxpce) and Claude.

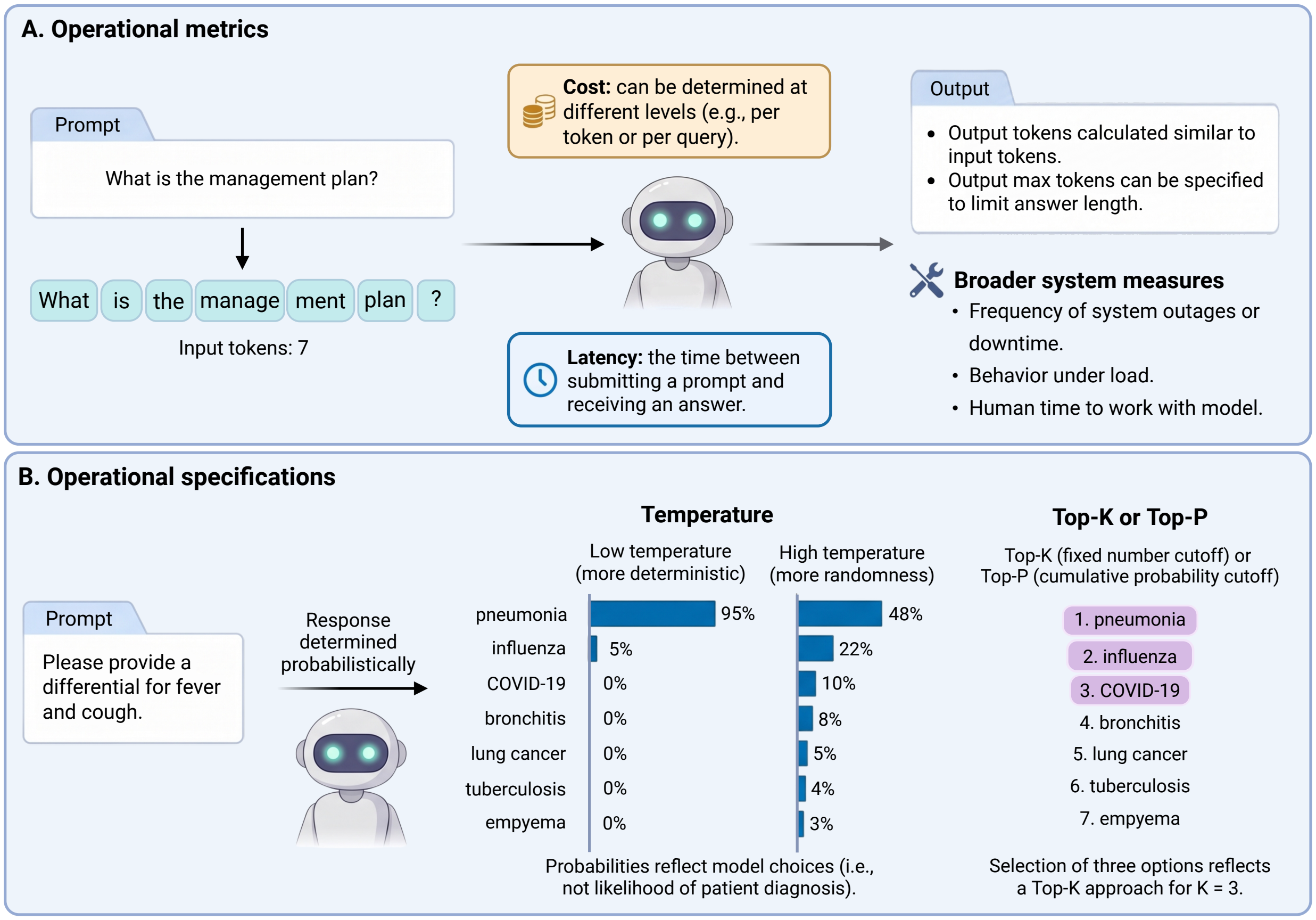

## Clinical application evaluation

Following capability evaluation, a key question is whether LLMs perform in the intended clinical context outside the structured framework of benchmarks (Figure 4).

### Accuracy of free text outputs

Many LLM tasks produce free text like patient message responses or discharge summaries. This text may reduce to categorical or numeric form such that traditional statistical methods can be used. Equally, the text may need to be considered more holistically beyond exact-match scoring.

#### *Natural Language Processing metrics*

Natural Language Processing (NLP) developed methods for quantifying free text accuracy over many decades. Overlap metrics like BLEU,[79] ROUGE,[80] and METEOR[81] score generated text by counting how many short word sequences (known as n-grams) are shared with reference text. They often require exact matches and may penalize a correct answer that is phrased differently. Embedding-based metrics like BERTScore[82] and cosine similarity[83] convert text into numerical vectors to capture meaning, then compare vectors with reference text. They are better at crediting paraphrasing but can remain insensitive to clinically important details like negations, numbers and named entities. Both approaches were designed for shorter, more constrained text, and are often inadequate for evaluating longer outputs typical of LLMs.

NLP also uses readability metrics to characterize text. The Flesch-Kincaid Grade Level[84] and Flesch Reading Ease[85] estimate reading difficulty from average sentence length and average number of syllables per word. They measure form, not accuracy or meaning, and a readable passage can still be wrong; they should be reported alongside an accuracy metric. They can be particularly helpful for assessing patient-facing text including education materials and plain-language summaries.

*Acceptable answers by human review*

Free-text outputs may be scored by human subject matter experts using structured scoring rubrics. This approach allows healthcare specific considerations like clinical safety, data privacy and ethical alignment to be explicitly evaluated. The rubrics are commonly determined using Delphi methodology.[46] Instruments such as QUEST also provide literature-derived dimensions for structured human review, replacing ad hoc rubrics that otherwise differ from study to study and prevent pooling of results.[86] They additionally make rater reliability measurable: with a defined instrument, inter-rater agreement can be reported and calibration performed. In one prospective deployment, physicians reviewing AI-generated discharge summaries completed a structured feedback instrument covering type of error, harm severity and likelihood.[87] Human review of model outputs within a clinical workflow is the gold standard in evaluation but is limited by scaling and may not reach consensus. Crowd-sourced evaluations have been proposed as a possible solution.[88]

Domain-specific frameworks are also readily available, given that evaluation criteria are often not portable across healthcare contexts. For example, principles for acute care transfer poorly to consumer wellness applications, which commonly involve no clinician in the loop and the user as the decision-maker. The SHARP (safety, helpfulness, accuracy, relevance, personalization) framework was developed specifically for consumer health and wellness LLMs.[89]

*Acceptable answers by LLM-as-a-judge review*

An LLM-as-a-judge (or multiple LLMs, which is referred to as LLM-as-a-jury) can assess outputs against predefined evaluation criteria and scoring frameworks.[90,91] This approach scales more easily than human review and is particularly useful when evaluating adherence to clinical rubrics or qualitative characteristics such as relevance. Its scalability also facilitates ongoing monitoring of AI systems deployed in clinical environments.[92] The LLM should be provided with clear instructions and, where appropriate, reference standards or exemplars. Several approaches can help ensure consistency and fairness. The LLM should remain fixed across a study to avoid introducing variation. It should also ideally differ in family from the evaluated models, since models from the same family can exhibit strong preference for their own outputs.[93] When using an LLM-as-a-judge, it is usually important to conduct a sample of the assessments with human review to check concordance; inter-human concordance can provide a baseline for this comparison.

Faithfulness metrics ask if generated text is supported by the source. They typically include two components: groundedness checks model statements trace back to supporting passages,[94] while hallucination rate reports the frequency of statements without support.[95] ACUEval is a tool that assesses the faithfulness of summarizations with an LLM-as-a-judge approach.[96] It breaks down summarizations into individual facts (referred to as atomic content units) and then compares them with source text. Similarly, VeriFact decomposes then verifies individual statements in LLM-generated documents like the hospital course in discharge summaries.[97]

Clinical trial methodology

Beyond evaluating LLM outputs, traditional clinical trial methodology can assess the impact of LLMs on clinical outcomes. This methodology includes randomized clinical trials, where randomization occurs at a provider or institution level. Common comparisons include physicians versus LLMs, physicians using versus physicians not using an LLM, and LLMs alone versus LLMs with physicians.[98–100] A challenge with any trial is balancing endpoint metrics that have clearer attribution to LLM use with metrics that have more substantial clinical meaning like mortality rate; the latter may have many confounding factors without appropriate controls.

**Figure 4: Clinical application evaluation.** Judging free text outputs (A) involves rubric development followed by human review and/or LLM-as-a-judge review; traditional Natural Language Processing metrics can also be used. Faithfulness (B) involves checking statements (also referred to as atomic content units) with reference text to determine groundedness and hallucination rate. Acronyms: CAP: community-acquired pneumonia; IV: intravenous; LLM: large language model. Figure created with BioRender (https://BioRender.com/pwo6gyh) and Claude.

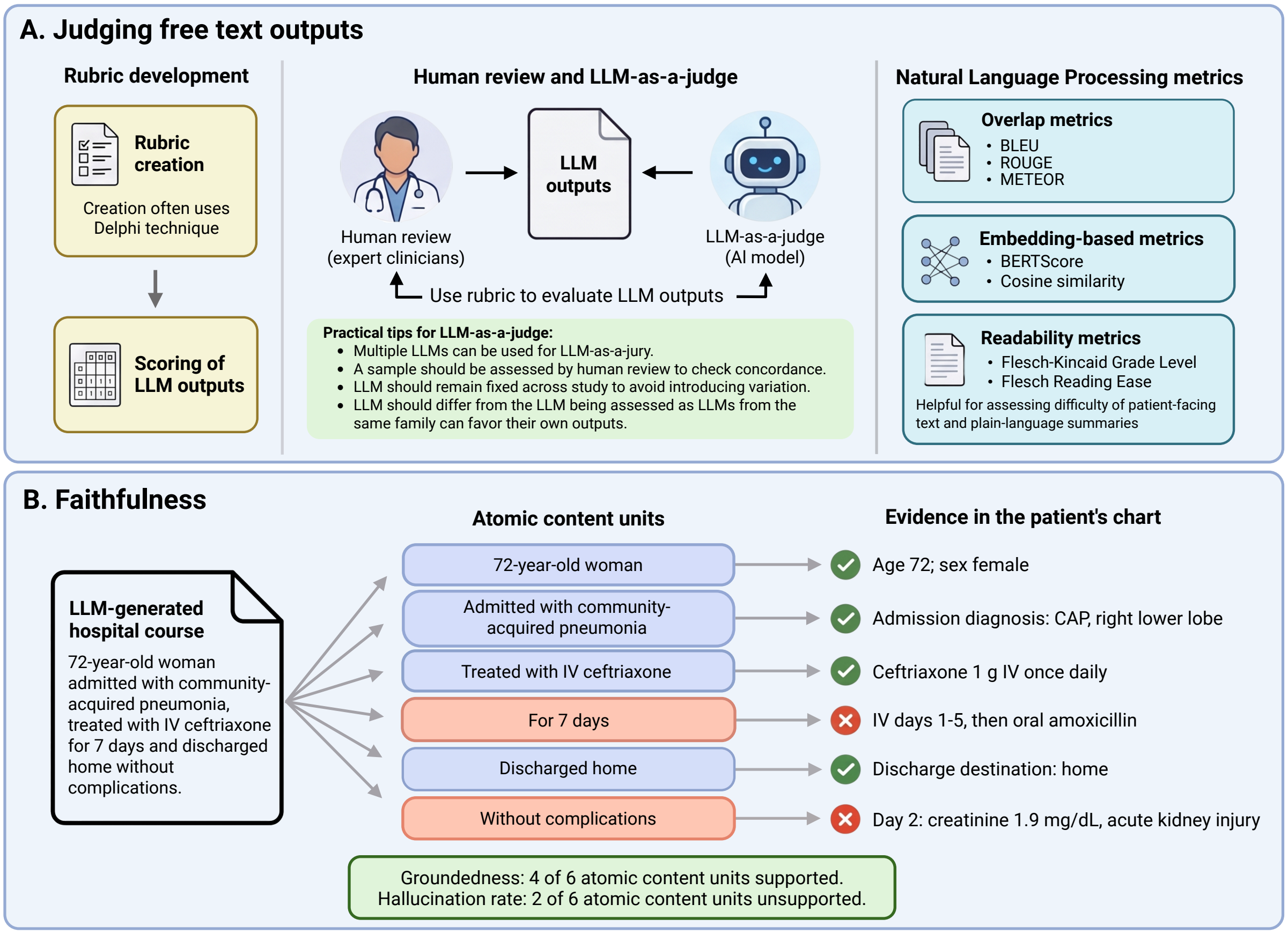

## Conclusion

LLMs have introduced new capabilities to medicine and provide promise for many more to follow. They require rigorous evaluation, which draws on existing study methodologies alongside newer methodologies to account for considerations like agentic systems and multi-turn conversations. This primer described four complementary areas: principles of study design, statistical methods, capability evaluation and clinical application evaluation. As LLM evaluations continue to evolve, there will likely be an increasing focus on real-world performance and clinical outcomes. Ultimately, rigorous healthcare LLM evaluation should not seek a single universal metric, but a transparent and reproducible framework for selecting complementary assessments that align with the research question being asked and the clinical decision that the model intends to support.